\documentclass[11pt,a4paper]{article}
\usepackage[hyperref]{acl2021}
\usepackage{times}
\usepackage{latexsym}

\usepackage{microtype}
\usepackage{amsmath}
\usepackage{amssymb}
\usepackage{url}
\usepackage{graphicx}
\usepackage{multirow}
\usepackage{CJKutf8}

\aclfinalcopy % Uncomment this line for the final submission
\newcommand{\Lagr}{\mathcal{L}}
\DeclareMathOperator*{\argmax}{arg\,max}
\DeclareMathOperator*{\argmin}{arg\,min}

\usepackage{enumitem}
\setlist[itemize]{leftmargin=*}

\title{BERTTune: Fine-Tuning Neural Machine Translation with BERTScore}

\author{Inigo Jauregi Unanue$^{\textbf{1,2}}$, Jacob Parnell$^\textbf{1,2}$, Massimo Piccardi$^\textbf{1}$\\
  $^1$University of Technology Sydney, NSW 2007, Australia\\
  $^2$RoZetta Technology, NSW 2000, Australia \\
  {\tt Inigo.Jauregi@rozettatechnology.com} \\
  {\tt Jacob.Parnell@rozettatechnology.com} \\
  {\tt Massimo.Piccardi@uts.edu.au}}

\date{}

\begin{document}
\maketitle
\begin{abstract}
Neural machine translation models are often biased toward the limited translation references seen during training. To amend this form of overfitting, in this paper we propose fine-tuning the models with a novel training objective based on the recently-proposed BERTScore evaluation metric. BERTScore is a scoring function based on contextual embeddings that overcomes the typical limitations of $n$-gram-based metrics (e.g. synonyms, paraphrases), allowing translations that are different from the references, yet close in the contextual embedding space, to be treated as substantially correct. To be able to use BERTScore as a training objective, we propose three approaches for generating \textit{soft predictions}, allowing the network to remain completely differentiable end-to-end. Experiments carried out over four, diverse language pairs have achieved improvements of up to $0.58$ pp ($3.28\%$) in BLEU score and up to $0.76$ pp ($0.98\%$) in BERTScore ($F_{BERT}$) when fine-tuning a strong baseline.
\end{abstract}

\section{Introduction}
\label{sec:intro}

Neural machine translation (NMT) has imposed itself as the most performing approach for automatic translation in a large variety of cases \cite{sutskever2014sequence,vaswani2017attention}. However, NMT models suffer from well-known limitations such as overfitting and moderate generalization, particularly when the training data are limited \cite{koehn2017six}. This mainly stems from the fact that NMT models have large capacity and are usually trained to maximize the likelihood of just a single reference sentence per source sentence, thus ignoring possible variations within the translation (e.g. synonyms, paraphrases) and potentially resulting in overfitting. A somewhat analogous problem affects evaluation, where metrics such as BLEU \cite{papineni2002bleu} only consider as correct the predicted $n$-grams that match exactly in the ground-truth sentence. 
In order to alleviate the $n$-gram matching issue during evaluation, Zhang et al. \shortcite{zhang2019bertscore} have recently proposed the BERTScore metric that measures the accuracy of a translation model in a contextual embedding space. In BERTScore, a pretrained language model (e.g. BERT \cite{devlin2019bert}) is first used to compute the contextual embeddings of the predicted sentence, $\langle \hat{\textbf{y}}_1, \dots, \hat{\textbf{y}}_k\rangle$, and the reference sentence, $\langle \textbf{y}_1, \dots, \textbf{y}_l\rangle$, with $k$ and $l$ word-pieces, respectively. Then, recall ($R_{BERT}$), precision ($P_{BERT}$), and F1 ($F_{BERT}$) scores are defined as cosine similarities between the normalized contextual embeddings. For example, the recall is defined as:
\begin{equation}
\label{eq:bert_recall}
R_{BERT} = \frac{1}{|l|} \sum_{y_i \in y} \max_{\hat{y}_j \in \hat{y}} \textbf{y}_i^T  \hat{\textbf{y}}_j
\end{equation}
\noindent where the $\max$ function acts as an alignment between each word in the reference sentence ($y$) and the words in the predicted sentence ($\hat{y}$). Conversely, $P_{BERT}$ aligns each word of the predicted sentence with the words of the reference sentence, and $F_{BERT}$ is the usual geometric mean of precision and recall. Note that with this scoring function a candidate and reference sentences with similar embeddings will be assigned a high score even if they differ completely in terms of categorical words.  Zhang et al. \shortcite{zhang2019bertscore} have shown that this evaluation metric has very high correlation with the human judgment.

In this work, we propose using BERTScore as an objective function for model fine-tuning. Our rationale is that BERTScore is a sentence-level objective that may be able to refine the performance of NMT models trained with the conventional, token-level log-likelihood. However, in order to fine-tune the model with BERTScore as an objective, end-to-end differentiability needs to be ensured. While the BERTScore scoring function is based on word embeddings and is in itself differentiable, its input derives from categorical predictions (i.e. argmax or sampling), breaking the differentiability of the overall model. In this work, we solve this problem by generating \textit{soft predictions} during training with three different approaches. One of the approaches, based on the Gumbel-Softmax \cite{jang2017categorical}, also leverages sampling, allowing the model to benefit from a certain degree of \textit{exploration}. For immediacy, we refer to our approach as \textit{BERTTune}. The experimental results over four, diverse language pairs have shown improvements of up to $0.58$ pp ($3.28\%$) in BLEU score and up to $0.76$ pp ($0.98\%$) in BERTScore with respect to a contemporary baseline \cite{ott2019fairseq}. 
% In this paper, we have focused in training the model to maximize the $F_{BERT}$ score which account for both precision and recall. 
%Section \ref{sec:experiments} we present several experiments on three language pairs that show how BERTScore can further fine-tune state-of-the-art transformer based NMT baselines.

\section{Related Work}
\label{sec:related}

In recent years, various researchers have addressed the problem of overfitting in NMT models. This problem can be specially severe for neural models, given that, in principle, their large number of parameters could allow for a perfect memorization of the training set. For instance, Ma et al. \shortcite{ma2018-bag} have trained an NMT model using both a reference sentence and its bag-of-words vector as targets, assuming that the space of alternative, correct translations share similar bags-of-words. Others \cite{elbayad2018token, chousa2018training} have proposed smoothing the probability distribution generated by the decoder using the embedding distance between the predicted and target words, forcing the network to increase the probability of words other than the reference. Another line of work has proposed to explicitly predict word embeddings, using the cosine similarity with the target embedding as the reward function \cite{kumar2019mises, jauregi2019rewe}. %A cosine distance loss is generally used to reward the model when it predicts vectors that are similar to the pretrained embeddings. 

% I DON'T THINK IT'S NEEDED:
%To the best of our knowledge, BERTScore \cite{zhang2019bertscore} is the first scoring function to adopt contextual embeddings rather than conventional word embeddings.

Reinforcement learning-style training has also been used to alleviate overfitting \cite{ranzato2015sequence, edunov2018classical}. The use of beam search removes the exposure bias problem \cite{wiseman2016sequence}, and the use of sampling introduces some degree of exploration. In addition, these approaches allow using non-differentiable, sequence-level metrics as reward functions. However, in practice, approximating the expectation of the objective function with only one or a few samples results in models with high variance and convergence issues. 
%and thus, a straight-through estimator that does not suffer from this shortcoming may be more desirable.

Significant effort has also been recently dedicated to leveraging large, pretrained language models \cite{devlin2019bert,radford2018improving, peters2018deep} for improving the performance of NMT models. This includes using contextual word embeddings either as input features \cite{edunov2019pre} or for input augmentation \cite{yang2020towards,zhu2020incorporating}, and using a pretrained language model for initializing the weights of the encoder \cite{clinchant2019use}. Alternatively, Baziotis et al. \shortcite{baziotis2020language} have proposed using a pretrained language model as a prior, encouraging the network to generate probability distributions that have a high likelihood in the language model. In abstractive summarization, Li et al. \shortcite{li2019deep} have used BERTScore as reward in a deep reinforcement learning framework. In a similar vein, our work, too, aims to leverage pretrained language models for improving the NMT accuracy. However, to the best of our knowledge, ours is the first work to directly include a language model as a differentiable evaluation measure in the training objective. In this way, the NMT model is able to exploit the value of a pretrained language model while at the same time being fine-tuned over a task-specific evaluation metric.

\section{BERTScore Optimization}
\label{sec:bert}

Translation evaluation metrics, including BERTScore, typically require a predicted translation, $\langle \hat{y}_1, \dots, \hat{y}_k\rangle$, and at least one reference translation, $\langle y_1, \dots, y_l\rangle$, as inputs. At its turn, the predicted translation is typically obtained as a sequence of individual word (or token) predictions, using beam search or greedy decoding. We can express the predictions as:
\begin{equation}
\label{eq:argmax}
\hat{y}_j = \argmax_y p(y|x,\hat{y}_{j-1},\theta) \quad j=1,\dots,k
\end{equation}
\noindent where $x$ represents the source sentence and $\theta$ the model's parameters. During model training, it is common practice to use teacher forcing (i.e., use words from the reference sentence as $\hat{y}_{j-1}$) for efficiency and faster convergence. 

In brief, the computation of BERTScore works as follows: the scorer first converts the words in the predicted and reference sentences to corresponding static (i.e., non-contextual) word embeddings using the embedding matrix, $\textbf{E}$, stored in the pretrained language model. For the predicted sequence, we note this lookup as:
\begin{equation}
\label{eq:embedding_lookup_pred}
\textbf{e}_{\hat{y}_j} = \text{emb}_{LM} (\textbf{E},\hat{y}_{j}) \quad j=1,\dots,k
\end{equation}
The sequences of static embeddings for the predicted and reference sentences are then used as inputs into the language model to generate corresponding sequences of contextualized embeddings, $\langle \hat{\textbf{y}}_1, \dots, \hat{\textbf{y}}_k\rangle$ and $\langle \textbf{y}_1, \dots, \textbf{y}_k\rangle$, respectively, over which the BERTScore is finally computed. For our work, we have chosen to optimize the $F_{BERT}$ score as it balances precision and recall. For more details on the scoring function we refer the reader to \cite{zhang2019bertscore}.

\subsection{Soft predictions}
\label{ssec:sampling}

However, it is not possible to directly use the $F_{BERT}$ score as a training objective since the argmax function in (\ref{eq:argmax}) is discontinuous.
%and prevents backpropagation.
Therefore, in this work we propose replacing the hard decision of the argmax with ``soft predictions'' that retain differentiability.
%Applying the softmax over the logits output by the decoder, $\textbf{s}_j$, yields a sequence of probability vectors, $\textbf{p}_j$, of the same size as the target vocabulary:
%\begin{equation}
%\label{eq:softmax}
%\textbf{p}^i_{j} = %\frac{e^{\textbf{s}^i_j}}{\sum_{v=1}^{V} %e^{\textbf{s}^v_j}}
%\end{equation}
Let us note concisely the probability in (\ref{eq:argmax}) as $p^i_j$, where $i$ indexes a particular word in the $V$-sized vocabulary and $j$ refers to the decoding step, and the entire probability vector at step $j$ as $\textbf{p}_j$. Let us also note as $\textbf{e}^i$ the embedding of the $i$-th word in the embedding matrix of the pretrained language model, $\textbf{E}$. We then compute an ``expected embedding'' as follows:
\vspace{-6pt}
\begin{equation}
\label{eq:marginalization}
\bar{\textbf{e}}_{\hat{y}_j} = \mathbb{E}[\textbf{E}]_{\textbf{p}_j} = \sum_{i=1}^V p^i_j \textbf{e}^i
\end{equation}
\noindent In other terms, probabilities $\textbf{p}_j$ act as attention weights over the word embeddings in matrix $\textbf{E}$, and the resulting expected embedding, $\bar{\textbf{e}}_{\hat{y}_j}$, can be seen as a trade-off, or weighted average, between the embeddings of the words with highest probability. To be able to compute this expectation, the NMT model must share the same target vocabulary as the pretrained language model. Once the expected embeddings for the whole predicted sentence, $\langle \bar{\textbf{e}}_{\hat{y}_1},\dots, \bar{\textbf{e}}_{\hat{y}_k} \rangle$, are computed, they are input into the language model to obtain the corresponding sequence of predicted contextualized embeddings, and the $F_{BERT}$ score is computed. The fine-tuning loss is simply set as $\Lagr = - F_{BERT}$. During fine-tuning, only the parameters of the NMT model are optimized while those of the pretrained language model are kept unchanged.

%\begin{equation}
%\label{eq:loss}
%\Lagr = - F_{BERT}(\langle \bar{\textbf{e}}_{\hat{y}_1},\dots, \bar{\textbf{e}}_{\hat{y}_k} \rangle,\langle %\textbf{e}_{y_1},\dots, \textbf{e}_{y_k} \rangle)
%\end{equation}

\subsection{Sparse soft predictions}
\label{sssec:sparse}

A potential limitation of using the probability vectors to obtain the expected embeddings is that they are, a priori, dense, with several words in the vocabulary possibly receiving a probability significantly higher than zero. In this case, the expected embeddings risk losing a clear interpretation. While we could simply employ a softmax with temperature to sparsify the probability vectors, we propose exploring two more contemporary approaches:

\begin{itemize}
    \item \textbf{Sparsemax} \cite{martins2016softmax}: Sparsemax generates a Euclidean projection of the logits computed by the decoder (noted as vector $\textbf{s}_j$) onto the probability simplex, $\Delta^{V-1}$:
    \begin{equation}
    \label{eq:sparsemax}
     \textbf{p}^{SM}_{j} = \argmin_{\textbf{p}_j\in\Delta^{V-1}}||\textbf{p}_j - \textbf{s}_j||^2
    \end{equation}
    The larger the logits, the more likely it is that the resulting $\textbf{p}^{SM}_j$ vector will have a large number of components equal to zero. The sparsemax operator is fully differentiable.

    \item \textbf{Gumbel-Softmax} \cite{jang2017categorical,maddison2017}: The Gumbel-Softmax is a recent re-parametrization technique that allows sampling \textit{soft} categorical variables by transforming samples of a Gumbel distribution. The transformation includes a temperature parameter, $\tau$, that allows making the resulting soft variables more or less sparse. By noting a  sample from the Gumbel distribution as $g^i$, the Gumbel-Softmax can be expressed as:
    \vspace{-6pt}
    \begin{equation}
    \label{eq:gumbel-softmax}
    p^{i \hspace{2pt} GS}_j = \frac{\exp((\log p^i_j+g^i)/\tau)}
    {\sum_{v=1}^V \exp((\log p^v_j + g^v)/\tau)}
    \end{equation}
    where $p^{i \hspace{2pt} GS}_j$, $i=1,\dots,V$, are the components of the probability vector used in (\ref{eq:marginalization}). In the experiments, $\tau$ has been set to $0.1$ to enforce sparsity. In addition to obtaining more ``selective'' predictions, the Gumbel-Softmax leverages sampling, allowing the fine-tuning to avail of a certain degree of \textit{exploration}. The Gumbel-Softmax, too, is fully differentiable.
    
\end{itemize}

%{\color{red}For ease of reference, we have nicknamed the proposed approach \textit{BERTTune}.}

\section{Experiments}
\label{sec:experiments}

\subsection{Datasets}
\label{ssec:resources}

\begin{table*}[t]	
	\begin{center}
		\centering
		\resizebox{\textwidth}{!}{\begin{tabular}{|l|c c c|c c c|c c c|c c c|}
			\hline	\multirow{2}{*}{\textbf{Model}}&\multicolumn{3}{|c|}{\textbf{de-en}}&\multicolumn{3}{|c|}{\textbf{zh-en}}&\multicolumn{3}{|c|}{\textbf{en-tr}}&\multicolumn{3}{|c|}{\textbf{en-es}}\\
			\cline{2-13}	&BLEU&$F_{BERT}$&MS&BLEU&$F_{BERT}$&MS&BLEU&$F_{BERT}$&MS&BLEU&$F_{BERT}$&MS\\
			\hline
			Transformer NMT&33.61&77.56&52.86&18.28&68.04&34.81&17.68&76.55&18.3&37.80&79.31&45.76\\
			\hline	
			\hspace{4pt}$+$ BERTTune (DV)&33.58&77.90&53.4$^\dag$&\textbf{18.53}$^\dag$&\textbf{68.53}$^\dag$&\textbf{35.57}$^\dag$&17.81$^\dag$&76.57&18.19&37.36&79.30&\textbf{45.92}$^\dag$\\
			\hspace{4pt}$+$ BERTTune (SM)&33.39&77.88&53.27$^\dag$&18.09&68.48$^\dag$&35.18$^\dag$&17.52&76.55&18.09&37.70&79.27&45.89$^\dag$\\
			\hspace{4pt}$+$ BERTTune (GS)&\textbf{33.97}&\textbf{78.32}$^\dag$&\textbf{53.58}$^\dag$&18.39$^\dag$&68.45$^\dag$&35.33$^\dag$&\textbf{18.26}$^\dag$&\textbf{76.75}$^\dag$&\textbf{18.33}&\textbf{37.96}$^\dag$&\textbf{79.33}&45.84$^\dag$\\
			\hline
		\end{tabular}}
		\vspace{-3pt}
		\caption{Average BLEU, $F_{BERT}$ and MoverScore (MS) results over the test sets. ($\dag$) refers to statistically significant differences with respect to the baseline computed with a bootstrap significance test with a $p$-value $< 0.01$  \cite{dror2018hitch}. The bootstrap test was carried out at sentence level for $F_{BERT}$ and MS, and at corpus level for BLEU.}
		\vspace{-15pt}
	   \label{tab:results}
    \end{center}
    
\end{table*}

We have carried out multiple experiments over four, diverse language pairs, namely, German-English (de-en), Chinese-English (zh-en), English-Turkish (en-tr) and English-Spanish (en-es), using the datasets from the well-known IWSLT 2014 shared task\footnote{\href{https://wit3.fbk.eu/2014-01}{https://wit3.fbk.eu/2014-01}}, with $152$K, $156$K, $141$K and $172$K training sentences, respectively. Following Edunov et al. \shortcite{edunov2018classical}, in the de-en dataset we have used $7,000$ samples of the training data for validation, and \textit{tst2010}, \textit{tst2011}, \textit{tst2012}, \textit{dev2010} and \textit{dev2012} as the test set. For the other language pairs, we have used the validation and test sets provided by the shared task. More details about the preprocessing are given in Appendix A.

\subsection{Models and training}
\label{ssec:train_and_hp}

We have implemented the fine-tuning objective using the \textit{fairseq} translation toolkit\footnote{ \href{https://github.com/ijauregiCMCRC/fairseq-bert-loss}{https://github.com/ijauregiCMCRC/fairseq-bert-loss}}  \cite{ott2019fairseq}. The pretrained language models  for each language have been downloaded from Hugging Face \cite{wolf2020transformers}\footnote{\href{https://huggingface.co/models}{https://huggingface.co/models}}.
%, namely \texttt{bert-base-uncased} (en), \texttt{dbmdz/bert-base-german-uncased} (de), \texttt{bert-base-chinese} (zh), \texttt{dbmdz/ bert-base-turkish-uncased} (tr) and \texttt{dccuchile/bert-base-spanish-wwm- uncased} (es).
As baseline, we have trained a full NMT transformer until convergence on the validation set. With this model, we have been able to reproduce or exceed the challenging baselines used in \cite{zhang2019bertscore,xia2019tied,miculicich2018document,wu2020mixed}.
%Because $F_{BERT}$ is a sentence-level score and to avoid a cold-start, we initialize another identical model with the weights of the baseline and we fine-tune the network with the new loss (again until convergence in validation).
The fine-tuning with the $F_{BERT}$ loss has been carried out over the trained baseline model, again until convergence on the validation set. For efficient training, we have used teacher forcing in all our models. During inference, we have used beam search with beam size $5$ and length penalty $1$. As performance measures, we report the BLEU, $F_{BERT}$ and MoverScore (MS) \cite{zhao2019moverscore} results over the test sets averaged over three independent runs. Including BLEU and MS in the evaluation allows us to probe the models on metrics different from that used for training. Similarly to $F_{BERT}$, MS, too, is a contextual embedding distance-based metric, but it leverages soft alignments (many-to-one) rather than hard alignments between words in the candidate and reference sentences. To make the evaluation more probing, for MS we have used different pretrained language models from those used with $F_{BERT}$. For more details on the models and hyperparameter selection, please refer to Appendix A.

\begin{figure}[t]
	\centering
	\includegraphics[width=\linewidth]{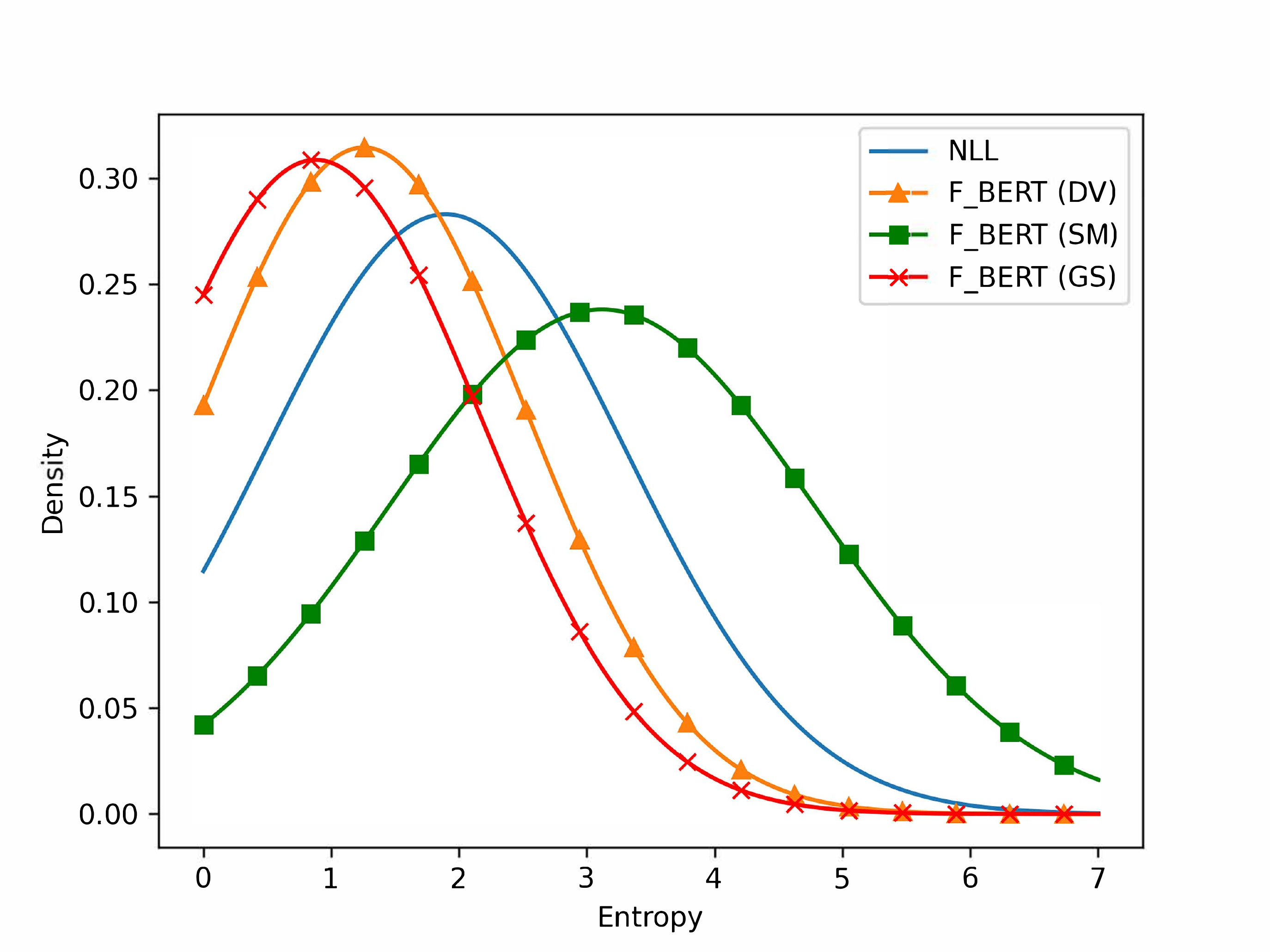}
	\vspace{-12pt}
	\caption{Entropy of the probability vectors generated by the different approaches over the de-en test set.}
	\vspace{-12pt}
	\label{fig:fig1}
\end{figure}

\subsection{Results}
\label{ssec:results}

Table \ref{tab:results} shows the main results over the respective test sets. As expected, fine-tuning the baseline with the proposed approach has generally helped improve the $F_{BERT}$ scores. However, Table \ref{tab:results} also shows that it has often led to improvements in BLEU score. In the majority of cases, the best results have been obtained with the Gumbel-Softmax (GS), with more marked improvements for de-en and en-tr ($+0.36$ pp BLEU and $+0.76$ pp $F_{BERT}$ and $+0.72$ pp MS for de-en, and $+0.58$ pp BLEU, $+0.20$ pp $F_{BERT}$  and  $+0.03$ pp MS for en-tr). Conversely, the dense vectors (DV) and sparsemax (SM) have not been as effective, with the exception of the dense vectors with the zh-en dataset ($+0.25$ pp BLEU, $+0.49$ pp $F_{BERT}$ and  $+0.54$ pp MS). This suggests that  the Gumbel-Softmax sampling may have played a useful role in exploring alternative word candidates. In fairness, none of the proposed approaches has obtained significant improvements with the en-es dataset. This might be due to the fact that the baseline is much stronger to start with, and thus more difficult to improve upon. In general, both the embedding-based metrics (i.e., $F_{BERT}$ and MS) have ranked the approaches in the same order, with the exception of the en-es dataset.
%Another possible cause could be a non-optimal selection of certain hyperparameters (e.g. $\tau$) or pretrained language model for the target language. Thus, in the near-future we are planning to explore more exhaustively the effectiveness of the proposed objective function with varying training data sizes, different sparsity levels (i.e. varying $\tau$) and different pretrained language models.

To provide further insights, similarly to Baziotis et al. \shortcite{baziotis2020language}, in Figure \ref{fig:fig1} we plot the distribution of the entropy of the probability vectors generated by the different approaches during inference over the de-en test set. Lower values of entropy correspond to sparser predictions. The plot shows that the models fine-tuned with the dense vectors and the Gumbel-Softmax have made test-time predictions that have been sparser on average than those of the baseline, with the Gumbel-Softmax being the sparsest, as expected. Conversely, and somehow unexpectedly, the model fine-tuned with the sparsemax has made predictions denser than the baseline's. We argue that this may be due to the scale of the logits that might have countered the aimed sparsification of the sparsemax operator. In all cases, the sparsity of the predictions seems to have positively correlated with the improvements in accuracy. For a qualitative analysis, Appendix B presents and discusses various comparative examples for different language pairs. % (de-en, zh-en).

\begin{figure}[t]
	\centering
	\includegraphics[width=\linewidth]{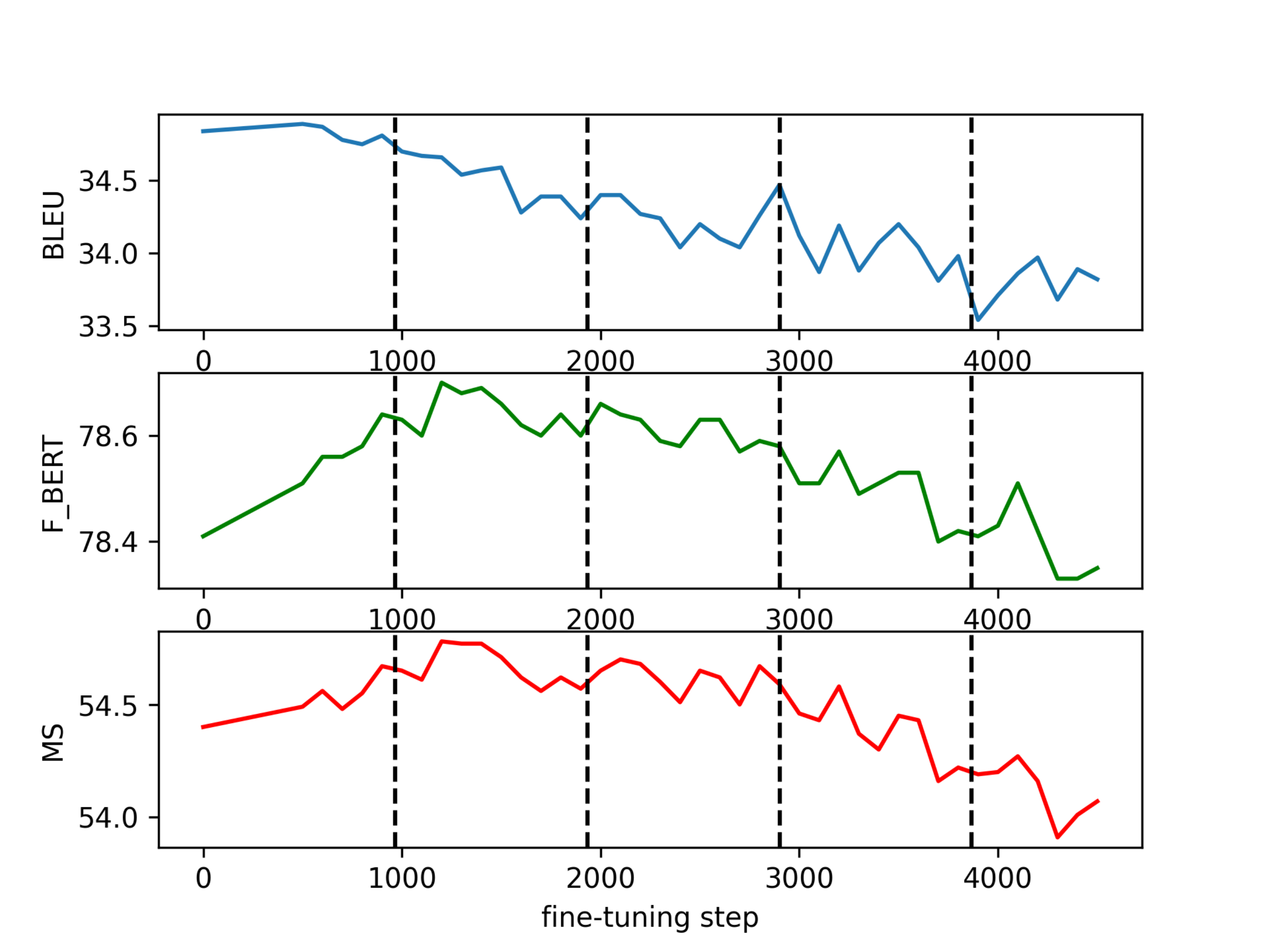}
	\vspace{-12pt}
	\caption{BLEU, $F_{BERT}$ and MS scores of the BERTTune (GS) model over the de-en validation set at different fine-tuning steps. Step 0 is the score of the baseline model, and the vertical dashed lines delimit the epochs.}
	\vspace{-12pt}
	\label{fig:fig2}
\end{figure}

Finally, Figure \ref{fig:fig2} shows the effect of the proposed objective over the measured metrics on the de-en validation set at different fine-tuning steps. The plots show that the model rapidly improves the performance in $F_{BERT}$ and MS scores during the first epoch (steps $1-967$), peaking in the second epoch ($\approx$ step $1,200$). After that, the performance of the model starts dropping, getting back to the baseline levels in epoch 4. 
%This is an interesting phenomenon that requires more analysis, which we will carry out as future work.
This suggests that training can be limited to a few epochs only, to prevent overfitting.
On the other hand, the plots also show a trade-off between the metrics, as the model's improvements in $F_{BERT}$ and MS come at cost of a decrease in BLEU. However, this phenomenon has not been  visible on the test set, where all the fine-tuned models have outperformed the baseline also in BLEU score. This suggests that for this dataset the distributions of the training and test sets may be more alike. 

\section{Conclusion}
\label{sec:conclusion}

In this work, we have proposed fine-tuning NMT models with BERTScore, a recently proposed word embedding-based evaluation metric aimed to overcome the typical limitations of $n$-gram matching. To be able to use BERTScore as an objective function while keeping the model end-to-end differentiable, we have proposed generating \textit{soft} predictions with differentiable operators such as the sparsemax and the Gumbel-Softmax. The experimental results over four language pairs have showed that the proposed approach -- nicknamed \textit{BERTTune} -- has been able to achieve statistically significant improvements in BLEU, $F_{BERT}$ and MS scores over a strong baseline.
As future work, we intend to explore the impact of key factors such as the dataset size, the sparsity degree of the predictions and the choice of different pretrained language models, and we also plan to evaluate the use of beam search/sequential sampling during training to leverage further exploration of candidate translations.

\bibliographystyle{acl_natbib}
\bibliography{acl2021}

\appendix

\newpage 

\phantom{blabla}

\newpage

\section*{Appendix A: Preprocessing and hyperparameters}
\label{appendix_a}

This appendix provides detailed information about the preprocessing of the datasets and the hyperparameter selection to facilitate the reproducibility of the experiments. All the code will be released after the anonymity period.

As part of the preprocessing of the datasets, all sentences have been tokenized and lowercased. The source languages have been tokenized with the Moses tokenizer\footnote{\href{https://github.com/moses-smt/mosesdecoder}{https://github.com/moses-smt/mosesdecoder}}, except for Chinese that has been tokenized using Jieba\footnote{\href{https://github.com/fxsjy/jieba}{https://github.com/fxsjy/jieba}}. The target languages have instead been tokenized with the tokenizer learned by the pretrained language model. As language models for BERTScore, we have used \texttt{bert-base-uncased} (en), \texttt{dbmdz/ bert-base-turkish-uncased} (tr) and \texttt{dccuchile/bert-base-spanish-wwm- uncased} (es) from Hugging Face. As language model for the MoverScore, we have used the suggested language model for English\footnote{\href{https://github.com/AIPHES/emnlp19-moverscore/releases/download/0.6/MNLI\_BERT.zip}{https://github.com/AIPHES/emnlp19-moverscore/releases/download/0.6/MNLI\_BERT.zip}},  \texttt{dbmdz/distilbert-base-turkish- cased} for Turkish and \texttt{mrm8488/distill-bert-base-spa nish -wwm-cased-fine tuned-spa-squad2-es} for Spanish, the last two from Huggingface. The few sentences longer than 175 tokens have been removed from all datasets as in the original fairseq preprocessing script. Additionally, further tokenization at subword level has been performed over the source languages using byte-pair encoding (BPE) \cite{sennrich2016neural} with $32,000$ merge operations. An important step in the preprocessing has been to force the decoder and the language model to share the same vocabulary. Therefore, we have assigned the decoder with the vocabulary from the selected pretrained language model, ensuring that both used identical \texttt{bos}, \texttt{eos}, \texttt{pad} and \texttt{unk} tokens.

For training a strong transformer baseline, we have followed the recommendations in fairseq\footnote{\href{https://fairseq.readthedocs.io/en/latest/index.html}{https://fairseq.readthedocs.io/en/latest/index.html}}. The architecture is the predefined \texttt{transformer\_iwslt\_de\_en} architecture ($79$M parameters) with word embedding and hidden vector dimension size of $512$, and $6$ transformer layers. We have set the training batch size to $4,096$ tokens, the dropout rate to $0.3$ and the \texttt{clip\_norm} gradient clipping parameter to $0.0$. The objective function is the label-smoothed negative log-likelihood, with the smoothing factor set to $0.1$. We have used the Adam optimizer \cite{kingma2014adam} with a $5e-4$ learning rate and beta values $\beta_1=0.9$ and $\beta_2=0.98$. We have set the warm-up steps to $4,000$ with an initial learning rate of $1e-7$. During training, we have reduced the learning rate with an inverse square-root scheduler and the weight decay set to $0.001$. We have trained the model until convergence of the BLEU score on the validation set, with checkpoints at each epoch and patience set to $3$, or until the learning rate dropped below $1e-9$.

For fine-tuning with $F_{BERT}$, we have initialized the transformer models with the trained weights of the baseline. We have kept all hyperparameters identical, except for the learning rate which has been reduced by an order of magnitude to $5e-5$, following common fine-tuning strategies. The models have been fine-tuned until convergence over the validation set, with patience set to $3$. Since the changes have only involved the training objective, the number of trainable parameters has remained exactly the same ($79$M). At test time, we have used beam search decoding with beam size $5$ and length penalty $1$.
 
For all the experiments we have used an NVIDIA Quadro P5000 GPU card with $16$ GB of memory. 

\section*{Appendix B: Translation examples}
\label{appendix_b}

%This appendix shows few translation examples and the predictions made by each model in order to help to understand the kind of improvements BERTTune can make.

This appendix shows a few translation examples from the de-en and zh-en language pairs to provide further insights into the behavior of the different models.

The example in Table \ref{tab:trans_example_1} shows that only the BERTTune model with the Gumbel-Softmax has been able to translate phrases such as \textit{at the moment} and \textit{it was as if / it was like}. This model seems to have been able to capture the exact meaning of the source German sentence, even though it has translated it with a slightly different wording (note that the Gumbel-Softmax fine-tuning explores a larger variety of predictions). The other BERTTune models, too, have translated this sentence better than the baseline. 

In the example in Table \ref{tab:trans_example_2}, the baseline has not been able to correctly pick the name of the artist (\textit{bono}, lowercased from \textit{Bono}), choosing instead word \textit{bonobos} (primates). All the BERTTune models have instead made the correct prediction. In this example, it is possible that the BERTTune models have benefited from the fine-tuning with a pretrained language model: word \textit{bono} might not have been present in the limited translation training data, but might have been encountered in the large unsupervised corpora used to train the language model. Another possibility is that they have simply used the copy mechanism more effectively.

In the example in Table \ref{tab:trans_example_3}, all the BERTTune models have correctly translated the phrase \textit{part of the national statistics}, while the baseline has incorrectly translated it as \textit{part of the world record}. In turn, the BERTTune models have translated the phrase \textit{in a decade or two} as \textit{in 10 or 20 years} which is a correct paraphrase, whereas the baseline has used the exact phrase as the reference. We also note that although both the baseline and BERTune translations have scored a BLEU score of $0.0$ in this case, the F$_{BERT}$ score has been able to differentiate between them, assigning a score of $72.36$ to the BERTTune translation and $72.10$ to the baseline. This also shows that small gains in F$_{BERT}$ score can correspond to significant improvements in translation quality.

Finally, in the example in Table \ref{tab:trans_example_4} only the BERTTune models with dense vectors and Gumbel-Softmax have been able to translate the beginning of the sentence (\textit{i was the guy beaten up}) with acceptable paraphrases (i.e. \textit{and i was the kind of person who had been beaten up / i was that guy who had been beaten}). Conversely, the baseline has translated the ending part of the sentence (\textit{until one teacher saved my life}) with a phrase of antithetical meaning (\textit{until a teacher turned me into this kind of life}).

\begin{table*}[h]	
	\begin{center}
		\centering
		\vspace{-2.5in} % NOTE THIS! To pull up.
		\resizebox{2\columnwidth}{!}{\begin{tabular}{|l l|}
				\hline
				%\textbf{Source}: &\makecell{Sentence in English our models prediction.\\ This is very important becaus there are serious\\ concersn about manythings in the world and}\\
				%\hline
				\textbf{Src}: & in dem moment war es , als ob ein filmregisseur einen bühnenwechsel verlangt hätte . \\
				\hline
				\textbf{Ref}: & \textcolor{green}{at that moment , it was as if a film director} called for a set change .\\
				\hline
				\textbf{Transformer NMT}: & the moment a film director would have asked a stager .\\
				\hline	
				\textbf{BERTTune (Dense vectors)} & and at that moment , a film director would have wanted a stage change .\\
				\hline	
				\textbf{BERTTune (Sparsemax)} & the moment a film director wanted a stage change .\\
				\hline	
				\textbf{BERTTune (Gumbel-Softmax)} & \textcolor{green}{at the moment , it was like a film director} would have wanted a stage change .\\
				\hline
		\end{tabular}}
		\caption{De-en translation example.}
		\label{tab:trans_example_1}
	\end{center}
%\end{table*}

\vspace{12pt}

%\begin{table*}[h]	
	\begin{center}
		\centering
		\resizebox{1.6\columnwidth}{!}{\begin{tabular}{|l l|}
				\hline
				%\textbf{Source}: &\makecell{Sentence in English our models prediction.\\ This is very important becaus there are serious\\ concersn about manythings in the world and}\\
				%\hline
				\textbf{Src}: & und interessanterweise ist bono auch ein ted prize gewinner . \\
				\hline
				\textbf{Ref}: & and interestingly enough , \textcolor{green}{bono} is also a ted prize winner .\\
				\hline
				\textbf{Transformer NMT}: & and interestingly , \textcolor{red}{bonobos} are also a ted prize winner .\\
				\hline	
				\textbf{BERTTune (Dense vectors)} & and interestingly , \textcolor{green}{bono} is also a ted prize winner .\\
				\hline	
				\textbf{BERTTune (Sparsemax)} & and interestingly , \textcolor{green}{bono} is also a ted prize winner .\\
				\hline	
				\textbf{BERTTune (Gumbel-Softmax)} & and interestingly , \textcolor{green}{bono} is also a ted prize winner .\\
				\hline
		\end{tabular}}
		\caption{Another de-en translation example.}
		\label{tab:trans_example_2}
	\end{center}
%\end{table*}

\vspace{12pt}

%\begin{table*}[h]	
	\begin{center}
		\centering
		\resizebox{2\columnwidth}{!}{\begin{tabular}{|l l|}
				\hline
				%\textbf{Source}: &\makecell{Sentence in English our models prediction.\\ This is very important becaus there are serious\\ concersn about manythings in the world and}\\
				%\hline
				\textbf{Src}: & \begin{CJK*}{UTF8}{gbsn}我 想 在 十年 或 二十年 内 ， 这 将 会 成为 国家 统计数据 的 一部分 。\end{CJK*} \\
				\hline
				\textbf{Ref}: & this is going to be , i think , within the next decade or two , \textcolor{green}{part of national statistics} .\\
				\hline
				\textbf{Transformer NMT}: & i think it ' s going to be \textcolor{red}{part of the world record} in a decade or two .\\
				\hline	
				\textbf{BERTTune (Dense vectors)} & and i think that in 10 or 20 years , this will be \textcolor{green}{part of the national statistics .}\\
				\hline	
				\textbf{BERTTune (Sparsemax)} & i think that in 10 or 20 years , this will be \textcolor{green}{part of the national statistics .}\\
				\hline	
				\textbf{BERTTune (Gumbel-Softmax)} & i think that in 10 or 20 years , this will be \textcolor{green}{part of the national statistics .}\\
				\hline
		\end{tabular}}
		\caption{Zh-en translation example.}
		\label{tab:trans_example_3}
	\end{center}
%\end{table*}

\vspace{6pt}

%\begin{table*}[h]	
	\begin{center}
	    \centering
		\resizebox{2\columnwidth}{!}{
		\begin{tabular}{|l l|}
				\hline
				%\textbf{Source}: &\makecell{Sentence in English our models prediction.\\ This is very important becaus there are serious\\ concersn about manythings in the world and}\\
				%\hline
				\textbf{Src}: & \begin{CJK*}{UTF8}{gbsn}我 是 那种 每周 在 男生宿舍 被 打 到 出血 的 那种 人 直到 一个 老师 把 我 从 这种 生活 中 解救出来。\end{CJK*} \\
				\hline
				\textbf{Ref}: & \textcolor{green}{i was the guy beaten up} bloody every week in the boys ' room , \textcolor{green}{until one teacher saved my life .}\\
				\hline
				\textbf{Transformer NMT}: & i was the one who was in the dorm room every week , and it wasn ' t  \textcolor{red}{until a teacher turned me into this kind of life} .\\
				\hline	
				\textbf{BERTTune (Dense vectors)} & \textcolor{green}{and i was the kind of person who had been beaten up} in the dorma every week until a teacher turned me out of this life .\\
				\hline	
				\textbf{BERTTune (Sparsemax)} & i ' m the kind of person who fell into his dorm room till a teacher turned me through this kind of life .\\
				\hline	
				\textbf{BERTTune (Gumbel-Softmax)} & \textcolor{green}{i was that guy who had been beaten} in his dorm room every week, until a teacher took me out of that life .\\
				\hline
		\end{tabular}}
		\caption{Another zh-en translation example.}
		\label{tab:trans_example_4}
	\end{center}
\end{table*}

\end{document}